\documentclass[10pt,journal,twocolumn]{IEEEtran}
\ifCLASSOPTIONcompsoc
  \usepackage[nocompress]{cite}
\else
  \usepackage{cite}
\fi

\ifCLASSINFOpdf
\else
\fi

\usepackage{times}
\usepackage{epsfig}
\usepackage{graphicx}
\usepackage{amsmath,amssymb}
\usepackage{eucal,bibspacing}
\usepackage{cite}
\usepackage{cs}
\usepackage{mathsymb}
\usepackage{textcomp}
\usepackage{verbatim}
\usepackage[super]{nth}
\usepackage{colortbl}
\usepackage{color}
\usepackage{xcolor}

\usepackage{multirow}
\usepackage{pbox}
\usepackage{wrapfig}
\usepackage{caption}
\usepackage[linesnumbered,algoruled,boxed,lined]{algorithm2e}

\let\savedalgorithm\algorithm
\let\savedendalgorithm\endalgorithm

\usepackage{algorithm}

\renewcommand{\etal}{\textit{et al. }}
\newcommand{\eg}{\textit{e.g., }}

\def\bf{{\boldsymbol f}}

\begin{document}

\title{Few-Shot Deep Adversarial Learning for Video-based Person Re-identification}

\author{Lin Wu, Yang Wang*, Hongzhi Yin, Meng Wang, Ling Shao
\IEEEcompsocitemizethanks{\IEEEcompsocthanksitem \protect\\ Lin Wu is with Key Laboratory of Knowledge Engineering with Big Data (Hefei University of Technology), Ministry of Education; School of Computer Science and Information Engineering, Hefei University of Technology, Hefei 230000, China. Email: jolin.lwu@gmail.com.
\protect\\ Yang Wang* (Corresponding author) is with School of Computer Science and Information Engineering, Hefei University of Technology, Hefei 230000, China. E-mail: yangwang@hfut.edu.cn. Yang Wang was supported by National Natural Science Foundation of China, under Grant No 61806035.
\protect\\ Hongzhi Yin is with The University of Queensland, St Lucia 4072, Australia. E-mail: db.hongzhi@gmail.com.
\protect\\ Meng Wang is with the School of Computer Science and Information Engineering, Hefei University of Technology, Hefei 230000, China. E-mail: eric.mengwang@gmail.com. Meng Wang was supported by National Natural Science Foundation of China, under Grant No 61432019, 61732008, and 61725203.
\protect\\ Ling Shao is with Inception Institute of Artificial Intelligence (IIAI), Abu Dhabi, United Arab Emirates. E-mail: ling.shao@ieee.org).
}
}

\IEEEtitleabstractindextext{%
\begin{abstract}
Video-based person re-identification (re-ID) refers to matching people across camera views from arbitrary unaligned video footages. Existing methods rely on supervision signals to optimise a projected space under which the distances between inter/intra-videos are maximised/minimised. However, this demands exhaustively labelling people across camera views, rendering them unable to be scaled in large networked cameras. Also, it is noticed that learning effective video representations with view invariance is not explicitly addressed for which features exhibit different distributions otherwise. Thus, matching videos for person re-ID demands flexible models to capture the dynamics in time-series observations and learn view-invariant representations with access to limited labeled training samples. In this paper, we propose a novel few-shot deep learning approach to video-based person re-ID, to learn comparable representations that are discriminative and view-invariant. The proposed method is developed on the variational recurrent neural networks (VRNNs) and trained adversarially to produce latent variables with temporal dependencies that are highly discriminative yet view-invariant in matching persons. Through extensive experiments conducted on three benchmark datasets, we empirically show the capability of our method in creating view-invariant temporal features and state-of-the-art performance achieved by our method.
\end{abstract}

\begin{IEEEkeywords}
Video-based person re-identification, Variational recurrent neural networks, Adversarial learning.
\end{IEEEkeywords}}

\maketitle

\IEEEdisplaynontitleabstractindextext

\IEEEpeerreviewmaketitle

\section{Introduction}\label{sec:intro}

\IEEEPARstart{A}{n} essential task in visual surveillance system is to automatically associate individuals across disjoint camera views, which is known as person re-identification (\textit{re-ID}). It has gained considerable popularity in video surveillance, multimedia, and security system by its prospect of searching persons of interest from a large amount of video sequences. Most existing approaches focus on matching still images represented by spatial visual appearance (shape, texture, and colour). Specifically, one matches a probe (or a query) person observed in one camera view against a set of gallery candidates captured by another disjoint view for generating a ranked list according to their matching distance or similarity. To this end, a body of methods have been developed to extract invariant features \cite{Farenzena2010Person,Gheissari2006Person,MidLevelFilter,Gray2008Viewpoint,eSDC} or learn discriminative matching models \cite{LADF,Multi-taskDistance,Kostinger2012Large,Zheng2013PAMI,Xiong2014Person,LOMOMetric,RankSVM}. However, people appearance is intrinsically limited due to the inevitable visual ambiguities and unreliability caused by appearance similarities among different people and appearance variations of the same person from significant cross-view changes in terms of poses, illumination, and cluttered background. This motivates the demand of seeking additionally helpful visual information sources for person re-ID.

On the other hand, videos or image sequences are often largely available from surveillance cameras, which inherently contain more information than independent images. One question raised is whether we can obtain more useful information from videos as opposed to still images? Videos are abundant and rich source of human motion and appearance information. For example, given sequences of images, temporal information related to a person's motion can be captured, which may disambiguate difficult cases that arise in the case of recognising the person in a different camera. However, working on videos creates new challenges such as dealing with video sequences of arbitrary length, and the difficulty of learning effective representations while disentangling nuisance factors caused by visual variations.


Most approaches to video-based person re-ID are based on supervised learning to optimise a discriminant distance metric under which minimised intra-video and maximised inter-video distances can be achieved \cite{Top-push,Video-person-ijcai16,VideoRanking}. They typically extract spatial-temporal features (\eg HOG3D \cite{HOG3D}) on each fragment from which videos are represented as a set of extracted features. To learn data-dependent high-level video features, \cite{RCNRe-id,RFA-net} use Long-Short Term Memory (LSTM) networks to aggregate frame-wise CNN features into video-level representations while mapping into hidden states with temporal dependency. However, videos are much higher dimensional entities, and it becomes increasingly difficult to do credit assignment on each frame selection to learn long-range relationship among them, unless we collect much more labeled data or do a lot of feature engineering (\eg computing the right kinds of flow features) to keep the dimensionality low. As a matter of fact, it is not realisable for person re-ID to collect massive labeled pairs of video sequences.

\paragraph{Challenges in Video Matching based Person Re-ID}
Matching video footages of pedestrians in realistic surveillance raises several challenges. First, it demands a faithful model with long distance dependency to map sequences into latent variables where both the dynamics of input patterns and effective representations can be learned simultaneously. Second, the model should be designed without demand on large amount of labeled training samples \cite{Unsupervised-Re-ID}. To this end, it desires a representation learning with access to few labeled data. Last but not the least, learning latent variables as video representations across camera views inherently exhibits distribution divergence, which should be eliminated to make the matching comparable \cite{Cross-GAN-Re-ID}. An illustration on feature distribution divergence is shown in Fig.\ref{fig:distribution_diver}. It is seen that the initial probability distributions of latent variables across views are very arbitrary (depicted in Fig.\ref{fig:distribution_diver} (a)). Also, it is noted that the cluttering of two distributions is caused by the high-dimensional hidden vectors in two views, while applying one-dimensional projection onto the hidden vectors is not suitable for unary matching. Thus, it requires a unified-view approach for constructing rich latent variables, and allows comparable matching in cross-view setting.Fig. \ref{fig:distribution_diver} (b) shows that after cross-view adversarial learning, the features are transformed to be view-invariant and comparable for matching (see Fig.\ref{fig:distribution_diver} (c), as measured by the KL-divergence).


\paragraph{Our Method}
w
In this work, we propose few-shot deep adversarial neural networks to learn latent variables as representations for cross-view video-based person re-ID in the context of few labeled training video pairs. Video observations are modelled using variational recurrent neural networks (VRNNs) \cite{VRNNs} that use variational methods to produce latent representations while capturing the temporal dependency across time-steps. To achieve view-invariance, we perform adversarial training to produce the latent representations invariant across camera views. Specifically, the VRNNs contain variational auto-encoders that provide a class of latent variables to capture the input dynamics, all of which conditioned on the previous encoders via hidden states of an RNN. To promote the learned features view-invariant, the network is augmented with cross-view verification to update adversarially into the view changes, and it encourages view-invariant features to emerge in the course of the optimisation. The proposed approach is generic as it can be created atop any existing feed-forward architecture that is trainable by back-propagation. Meanwhile, the network is easy to be optimised by adding a gradient reversal layer \cite{DANN} that leaves the input unchanged during forward passing and reverse the gradient by multiplying it by a negative scalar during back-propagation.

The inputs to the model are high-level percepts for a pair of videos, extracted by applying a convolutional net, namely VGG-Net \cite{VGG} trained on ImageNet \cite{AlexNet}. These percepts are the states of last layers of rectified linear hidden states form a CNN, which are put through the VRNNs to capture the latent dependencies at different time-steps. However, the derived latent variables from VRNNs are less regulated and exhibit varied distributions caused by view changes \cite{LDAFisherVector}. To this end, we propose adversarial learning which applies a cross-view verification to explicitly promote the view-invariant feature learning. In order to evaluate the learned representations, we qualitatively and quantitatively analyse the predictions and matching rates in people recognition made by the model.

\paragraph{Contributions} The major contributions of this paper are three-fold:
\begin{itemize}
\item We propose deep few-shot adversarial learning to produce effective video representations for video-based person re-ID with few labeled training paired videos. The proposed model atop VRNNs \cite{VRNNs} is able to map video sequences into latent variables which serve as video representations by capturing the temporal dynamics.
\item Our approach addresses the distribution disparity of learned deep features by performing adversarial training \cite{DANN} to ensure the view-invariance. Also, the algorithm is based on few-shot learning, and thus it is advantageous in generalisation capability to widen its application to large-scale networked cameras.
\item Extensive experiments are performed on three challenging benchmarks: iLIDS-VID \cite{VideoRanking}, PRID2011 \cite{PRID2011}, and MARS \cite{MARS} to show the notable improvement achieved by our method.
\end{itemize}

\section{Related Work}\label{sec:related}

\subsection{Person Re-identification}

The majority of approaches to person re-identification are image-based, which can be generally categorized into two categories. The first category \cite{Farenzena2010Person,MidLevelFilter,Gray2008Viewpoint,eSDC,What-and-where,Re-ID-CCA} employs invariant features that aims to extract features that are both discriminative and invariant against dramatic visual changes across views. The second stream is metric learning based method which are working by extracting features for each image, and then learning a metric where the training data have strong inter-class differences and intra-class similarities \cite{Pedagadi2013Local,LADF,KISSME,NullSpace-Reid,Deep-Embed,Yang-TIP15}. These approaches only consider one-shot image per person per view, which is inherently weak when multi-shot are available, due to the intrinsically ambiguous and noisy people appearance, and large cross-view appearance variations.

The use of video in many realistic scenarios indicates that multiple images can be exploited to improve the matching performance. Multi-shot approaches in person re-identification \cite{Farenzena2010Person,Gheissari2006Person,VideoRanking,HumanMatch,LOMOMetric,SpatioTempPRL} use multiple images of a person to extract the appearance descriptors to model person appearance.  For these methods, multiple images from a sequence are used to either enhance local image region/patch spatial feature descriptions \cite{Farenzena2010Person,Gheissari2006Person,HumanMatch,LOMOMetric} or to extract additional appearance information such as temporal change statistics \cite{SpatioTempPRL,Vehicle-Re-ID}. These methods, however, deal with multiple images independently whereas in the video-based person re-identification problem, a video contains more information than independent images, \eg underlying dynamics of a moving person and temporal evolution.

Recently, a number of frameworks are developed to address the problem of person re-ID in the video setting \cite{Video-person-ijcai16,VideoPerson,VideoRanking,Top-push}. The Dynamic Time Warping (DTW) model \cite{DTW}, which is a common sequence matching algorithm widely used in pattern recognition, has been applied into video-based person re-ID \cite{DTW-PersonREID}. Wang \etal \cite{VideoRanking,VideoRanking-pami} partially solve this problem by formulating a discriminative video ranking (DVR) model using the space-time HOG3D feature \cite{HOG3D}. In \cite{VideoPerson}, both spatial and temporal alignment is considered to generate a body-action model from which Fisher vectors are learned and extracted from individual body-action units, known as STFV3D. However, these approaches presumably suppose all image sequences are synchronised, whereas they are inapplicable in practice due to different actions taken by different people. An unsupervised approach is introduced by \cite{Video-person-matching} where a spatial-temporal person representation is introduced by encoding multi-scale spatial-temporal dynamics (including histograms of oriented 3D spatiotemporal gradient and spatiotemporal pyramids) from the unaligned image sequences.

A supervised top-push distance learning model (TDL) \cite{Top-push} is proposed to enforce top-push constraint \cite{Top-rank} into the optimisation on top-rank matching in person re-ID. It uses video-based representations (HOG3D, colour histograms and LBP) to maximise the inter-class margin between different persons, and likewise in \cite{Video-person-ijcai16}. Nonetheless, these pipelines perform feature extraction and metric learning separately in which low-level features are very generic to determine which of them are useful in matching process. Meanwhile, the process of distance metric learning is principled on individual examples (\eg triplets), which carries little information about neighbourhood structure, and thus not generalised across data sets.

With remarkable success achieved by deep neural networks (DNNs) in visual recognition, person re-ID has witnessed great progress by applying DNNs to learn ranking functions based on pairs \cite{PersonNet,JointRe-id,FPNN,DeepReID} or triplets of images \cite{DeepRanking}. These methods, which use network architectures such as the ``Siamese network'', learn a direct mapping from the raw image pixels to a feature space where diverse images from the same person are close, while images from different person are widely separated. Another DNN-based approach to re-ID, uses an auto-encoder to learn an invariant colour feature, whilst ignoring spatial features \cite{InvariantColor}, which turn out to be very crucial in matching pedestrians \cite{Correspondence,3D-PersonVLAD,Multiregion,Wu-TCYB}. However, existing architectures do not exploit any form of temporal information, and thus not applicable into video-based person re-ID. In order to introduce temporal signals into a DNN, McLaughlin \etal present a recurrent neural network for video-based person re-identification \cite{RCNRe-id}. It combines recurrence and temporal-pooling of appearance data with representation learning to yield an invariant representation for each person's video sequence. Wu \etal \cite{Wu-TMM} deliver an end-to-end approach to learn spatial-temporal features and corresponding similarity metric given a pair of time series.

\subsection{Deep Generative Models for Videos}

Ranzato \etal \cite{RanzatoVideoBaseline} proposed a generative model for videos by using a recurrent neural network to predict the next frame or interpolate between frames. In this work, the author quantise image patches into a large dictionary and train the model to predict the identity of the target patch. However, it introduces an arbitrary dictionary size and altogether removes the idea of patches being similar to dissimilar to one other. A recent model of using Encoder-Decoder LSTM to learn representations of video sequences is proposed by Srivastava \cite{VideoLSTM} where Encoder LSTM maps an input sequence into a fixed length representation, and Decoder LSTM produces the future sequence. Nonetheless, existing studies are dependent on Recurrent Neural Networks (RNNs) where the functions are deterministic and cannot capture variability in the input space. In this work, we study the visual variability in video sequences and capture the variability by developing a recurrent gaussian process model. The family of Recurrent Gaussian Process (RGP) models are defined by Lincoln \etal \cite{RecurrentGP}, which, similarly to RNNs, are able to learn temporal representations from data. Also, the authors propose a novel RGP model with a latent auto-regressive structure where the intractability brought by the recurrent GP priors are tackled with a variational approximation approach. While our method is similar to RGP \cite{RecurrentGP} by employing auto-regressive structure as latent states, this work addresses the distribution problem of latent variables effectively such that it provides the best approximation to the true posterior which is not resolved in \cite{RecurrentGP}. In contrast to the model of \cite{RecurrentGP} that needs expensive iterative inference scheme to continuous latent variables with intractable posterior distributions, we perform an efficient inference and a learning algorithm that even works in the intractable case. Our inference model based on the variational Bayes \cite{AEV,CYC-DGH} is able to re-parameterise the variational lower bound, which can be straightforwardly optimised using standard stochastic gradient descent techniques.

\subsection{Few-Shot Learning}

Few-shot learning is to learn novel concepts from very few examples \cite{low-shot}. It typically demands an effective representation learning pipeline that have good generalisation ability. Generative models can be used to generate additional examples so as to improve the learner's performance on novel classes \cite{one-shot-composition,low-shot}. Intuitively, the challenge is that these training examples capture very little of the category's intra-class variations. For instance, if the category is a particular pedestrian, then we may only have examples of the person's frontal view in one camera, and one of he/she in back view. Amongst representation learning approaches, metric learning such as triplet loss \cite{FaceNet,Web-scale-face,DeepRanking,DingLWC15} or Siamese networks \cite{Dim-reduction2006,one-shot-siamese,PersonNet,DeepReID} have been used to automatically learn feature representations where objects of the same class are put closer together. However, these approaches cannot be directly applied into video-based person re-ID because they do not explicitly address the cross-view variations which may lead to the feature distribution divergence.

\section{Few-Shot Deep Adversarial Neural Networks for Cross-View Video-based Person Re-Identification}\label{sec:approach}

In this section, we present a deep adversarial learning to learn video representations for person re-ID in a few-shot learning. In the case of video-based person re-ID, we need to learn from a few sequence examples regarding each person to produce discriminative representations. This motivates the setting we are interested in: ``few-shot" learning, which consists of learning a persons/class from little labelled sequence examples. While deep learning requires large training datesets to update the set of parameters, we develop our model by using the variational recurrent neural networks (VRNNs) with continuous latent random variables to model the video sequences, which allow us to perform efficient inference and learning. Then the learned latent temporal representations are put through adversarial training to make them view-invariant. In the following, we first formally describe the problem setting and definitions in Section \ref{ssec:problem-formulation}. Then, we present the VRNNs and the principle of adversarial learning in Section \ref{ssec:VRNNs} and Section \ref{ssec:adversarial-learning}, respectively. Section \ref{ssec:optimisation} contains the optimisation and Section \ref{ssec:inference} details the inference for testing.

\subsection{Problem Formulation}\label{ssec:problem-formulation}
Given a set of variable-length video sequences captured in a network of cameras with $N$ video observations $\{X^i=[x_t^i]_{t=1}^{T^i}\}_{i=1}^{N}$ where $X^i$ is a person video containing frames $x_t^i \in \mathbb{R}^D$. In the case of person re-ID, a video sequence regarding a person appearing in one camera is known as the \textit{probe video}, and the re-identification is to find the correct match for the probe from a set of \textit{gallery videos}. In this setting, we further assume each video sequence in the probe camera $\{X_p^i\}_{i=1}^{N_p}$ and its correspondence $\{X_g^i\}_{i=1}^{N_g}$ in the gallery set come with $L$ labels $y_i\in \{0,1\}^L$ (each person corresponds to a class). Without loss of generality, $N_p$ does not have to equal to $N_g$. Hence, each training pair is consisted of a randomly selected probe sequence $X^i_p$ and its correspondence $X_g$ regarding the same person $i$ with the label $y_i \in L$ where $L$ denotes the total number of persons. In overall, the training objective of our framework is to jointly achieve two goals: predict the label for each coupled probe and gallery video pair; and impose adversarial learning to regularise the learned representations to be view-invariant.

\subsection{Variational Recurrent Neural Networks (VRNNs)}\label{ssec:VRNNs}
To model the temporal dependencies between the latent random variable across time steps, we employ the variational recurrent neural networks (VRNNs) \cite{VRNNs}, which contains a variational auto-encoders \cite{AEV} at each time step. These encoders are conditioned on previous ones via the hidden state $h_{t-1}$ of an RNN such as an LSTM \cite{lstm1997}. Thus, for each time-step of frame $x_t^i$, a latent random variable $z_t^i$ can be inferred as

\begin{equation}\label{eq:latent-z}
  z_t^i | x_t^i \sim \mathcal{N}(\mu_{z,t}, diag(\sigma_{z,t})), [\mu_{z,t}, \sigma_{z,t}]=\phi_{\tau}^{Enc} \left( \phi_{\tau}^x (x_t^i), h_{t-1}\right)
\end{equation}
with the prior $z_t^i \sim \mathcal{N}(\mu_{0,t}, diag(\sigma_{0,t}))$ where $[\mu_{0,t},\sigma_{0,t}]=\phi_{\tau}^{prior} (h_{t-1})$. All $\mu_{\ast, t}, \sigma_{\ast, t}$ denote parameters of generating a distribution, and $\phi_{\tau}^{\ast}$ can be any highly flexible function such as deep neural networks. Then for each $z_t^i$, the data $x_t^i$ can be generated via

\begin{equation}
x_t^i | z_t^i \sim  \mathcal{N}(\mu_{x,t}, diag(\sigma_{x,t})), [\mu_{x,t}, \sigma_{x,t}]=\phi_{\tau}^{Dec} \left( \phi_{\tau}^z (z_t^i), h_{t-1}\right)
\end{equation}
and learned by optimising the VRNN objective function:

\begin{equation}\label{eq:VRNN}\small
\begin{split}
& \mathcal{L}_V (x_t^i, \theta_e, \theta_g)\\
& = \mathbb{E}_{q_{\theta_e} (z^i_{\leq T^i} |x^i_{\leq T^i})} [ \sum_{t=1}^{T^i} ( -KL(q_{\theta_e} (z_t^i | x_{\leq t}^i, z_{\leq t}^i) || p(z_t^i|x_{<t}^i, z_{<t}^i)) \\
& + \log p_{\theta_g} (x_t^i | z_{\leq t}^i, x_{< t}^i) )  ]
\end{split}
\end{equation}
where $KL(Q||P)$ is Kullback-Leibler divergence between two distributions $Q$ and $P$. $q_{\theta_e} (z_t^i | x_{\leq t}^i, z_{\leq t}^i)$ is the inference model,  $p(z_t^i|x_{<t}^i, z_{<t}^i)$ is the prior, $p_{\theta_g} (x_t^i | z_{\leq t}^i, x_{< t}^i)$ is the generative model. $\theta_e$ and $\theta_g$ are parameters of the VRNN's encoder and decoder, respectively. Thus, for each frame sequence $X^i$, we use $\bar{z}^i \sim q_{\theta_e} (z_{T^i}^i | x_{\leq T^i}^i, z_{\leq T^i}^i)$ as the overall feature representations for the following classification task since it captures temporal latent dependencies across the time steps.
%

\subsection{Deep Adversarial Learning for View-Invariance}\label{ssec:adversarial-learning}

Recall that in the context of few-shot person re-ID there are limited labeled pairs of video sequences in training, we perform the training objective by jointly optimising two tasks: the classification on each sequence and the verification on the correspondence. Intuitively, imposing the classification loss on inputs is able to optimise discriminative classifiers regarding identities with relative similarity in the context of all training classes because the optimal similarity is probabilistically determined by respecting all training categories \cite{LDAFisherVector,Wu-TMM}. The verification loss is principled to regularise the classifier to be view-invariance.

Formally, let $G_y (\bar{z}_p^i; \theta_y)$ and $G_d(\bar{z}_g^i, \theta_d)$ represent the probe classifier (predict class labels $y_i\in \{0,1\}^L$ for $X^i_p$) and gallery classifier (predict class labels $d_i\in \{0,1\}^L$ for $X^i_g$) respectively with parameters $\theta_y$ and $\theta_d$ for a given paired input $[\bar{z}_p^i, \bar{z}_g^i]$. Here, $G_y(\cdot)$ and $G_d(\cdot)$ can be modelled using deep neural networks. For the ease of notations, we set $L=N_p=N_g$, which suggests each pair input is coupled with one person and the training progressively considers one person out of $L$ identities. Therefore, the classification loss towards $X_p^i$ and $X_g^i$ ($i\in L$) can be defined respectively as
\begin{equation}\label{eq:classification}
\begin{split}
&\mathcal{L}_y (X_p^i; \theta_y, \theta_e) = \mathcal{L}_B \left(G_y( V_e(X_p^i; \theta_e); \theta_y ),  y_i\right); \\
&\mathcal{L}_d (X_g^i; \theta_d, \theta_e) = \mathcal{L}_B \left(G_d( V_e(X_g^i; \theta_e); \theta_d ) , d_i\right);
\end{split}
\end{equation}
where $\mathcal{L}_B$ represents the categorical cross-entropy loss function, and $V_e(X^i_{\ast}; \theta_e)$ is the VRNN encoder that maps input $X^i_{\ast}$ into its latent representation $\bar{z}_{\ast}^i$. Hence, for notation convenience, we define the combined classification loss as
\begin{equation}\label{eq:com-classification}
\mathcal{L}_C (X^i_{\ast}; \theta_y, \theta_d, \theta_e)= \mathcal{L}_y (X_p^i; \theta_y, \theta_e) + \mathcal{L}_d (X_g^i; \theta_d, \theta_e)
\end{equation}

Combing the VRNN training and classification task can yield the following optimisation:
\begin{equation}\label{eq:VRNN-classification}
\min_{\theta_e, \theta_g, \theta_y, \theta_d} \frac{1}{L} \sum_{i=1}^L \left( \frac{1}{T^i} \mathcal{L}_V (X^i_{\ast}, \theta_e, \theta_g) + \mathcal{L}_C (X^i_{\ast}; \theta_y, \theta_d, \theta_e) \right).
\end{equation}

As we are aimed in achieving the representations $\bar{z}^i$ with view-invariance, we can adversarially train the above objective function by incorporating the verification loss. In other words, the verification loss is introduced as the regulariser of Eq. \eqref{eq:VRNN-classification}:
\begin{equation}\label{eq:verification}
\mathcal{L}_R (\theta_e) = \max_{\theta_y, \theta_d} \left( -\frac{1}{L} \sum_{i=1}^L \left(\mathcal{L}_d (X_p^i; \theta_d, \theta_e) + \mathcal{L}_y (X_g^i; \theta_y, \theta_e) \right)\right).
\end{equation}

Jointly combining the optimisation problems in Eq. \eqref{eq:VRNN-classification} and Eq. \eqref{eq:verification} leads to our objective function, which can be mathematically express as:
\begin{equation}\label{eq:objective}
\small
\begin{split}
&\mathbb{E}(\theta_e, \theta_g, \theta_y, \theta_d)\\
&= \frac{1}{L} \sum_{i=1}^L \left( \frac{1}{T^i} \mathcal{L}_V (X^i_{\ast}, \theta_e, \theta_g) + \mathcal{L}_C (X^i_{\ast}; \theta_y, \theta_d, \theta_e) \right) \\
& + \lambda \mathcal{L}_R (\theta_e),
\end{split}
\end{equation}
where $\lambda$ is a tuning hyper-parameter to balance the trade-off between optimising on making view-invariant representations and optimising the classification accuracy.


\subsection{Optimisation}\label{ssec:optimisation}

The optimisation of Eq. \eqref{eq:objective} involves minimisation on some parameters, and maximisation on others, i.e., we iteratively solve the following problems by finding the saddle points $\hat{\theta}_e$, $\hat{\theta}_g$, $\hat{\theta}_y$, $\hat{\theta}_d$ such that:
\begin{equation}\label{eq:saddle}
\begin{split}
     & (\hat{\theta}_e, \hat{\theta}_g, \hat{\theta}_y, \hat{\theta}_d) =\arg \min_{\theta_e, \theta_g, \theta_y, \theta_d} \mathbb{E}(\theta_e, \theta_g, \hat{\theta}_y, \hat{\theta}_d); \\
     & (\hat{\theta}_d, \hat{\theta}_y)= \arg \max_{\theta_d, \theta_y} \mathbb{E}(\hat{\theta}_e, \hat{\theta}_g, \theta_y, \theta_d)
\end{split}
\end{equation}

This problem can be tackled with a simple stochastic gradient procedure, which is to make updates in the opposite direction of the gradient of Eq. \eqref{eq:objective} for the minimisation of parameters, and in the direction of the gradient for the maximisation of parameters. In practice, stochastic estimates of the gradient can be computed by using a subset of the training examples to calculate their averages. Thus, the gradient updates can be calculated as:
\begin{equation}\label{eq:gradient}
\begin{split}
     & \theta_e \leftarrow \theta_e -\eta (\frac{\partial \mathcal{L}_V}{\partial \theta_e} + \frac{\partial \mathcal{L}_C}{\partial \theta_e} - \lambda \frac{\partial \mathcal{L}_R}{\partial \theta_e}) \\
     & \theta_g \leftarrow \theta_g -\eta (\frac{\partial \mathcal{L}_V}{\partial \theta_g}) \\
     & \theta_y \leftarrow \theta_y -\eta ( \frac{\partial \mathcal{L}_y}{\partial \theta_y} -\lambda \frac{\partial \mathcal{L}_R}{\partial \theta_y} )\\
     & \theta_d \leftarrow \theta_d - \eta ( \frac{\partial \mathcal{L}_d}{\partial \theta_d} - \lambda \frac{\partial \mathcal{L}_R}{\partial \theta_d} )
\end{split}
\end{equation}

We use stochastic gradient decent (SGD) to update $\theta_g$. For the other parameters, we use SGD and gradient reversal layer \cite{DANN} to update a feed-forward deep networks that comprises feature extractor (VRNN's encoder) fed into the classification task ($\mathcal{L}_y$ and $\mathcal{L}_d$) and the cross-view verification loss. The role of gradient reversal is to make the gradients from the classification and view difference subtracted instead of being summed. This ensures the classification loss is maximised while the feature representations are view-invariant. Thus, the resulting feature representations can capture temporal dependencies (due to the VRNN objective function $\mathcal{L}_V$) and also cross-view invariance (due to the regressor of $\mathcal{L}_R$). The optimisation is depicted in Fig.\ref{fig:adversarial-train}.

\textit{Why should this framework learn good video features for video-based person re-ID?} The learned latent variables $z_i$ are not comparable across videos in cross-view setting which exhibit view variability and distributions. The formalism of cross-view verification offers a systematic mechanism of reducing view variations through generalising the classifiers where the loss function is to penalise the differences between the classifiers undr each view. Thus, this can draw connections between this loss and regularisation of feature activations.

\subsection{Inference and Complexity Analysis}\label{ssec:inference}

Once the training is accomplished, the inference model $q_{\theta_e} (\cdot)$ can be used to produce the feature representations for a video sequence following the equation: $z_t | x_t \sim \mathcal{N} (\mu_{z,t}, diag(\sigma_{z,t}))$, where $\mu_{z,t}$ and $\sigma_{z,t}$ denote the parameters of a Gaussian $\mathcal{N}(\cdot)$ whose mean $\mu$ and variance $\sigma^2$ are the output of a non-linear function of $x_t$, that is, $[\mu,\sigma]=\phi_r (\phi_r^x (x_t), h_{t-1})$. Given two unknown sequences in the test ${X}_p$ and ${X}_g$, we are able to obtain the latent variables $z_p$ and $z_g$ as their respective video representations to compute their similarity value via the element-wise inner product between the latent variables ${z}_p$ and ${z}_g$ \cite{Wu-TMM}, that is, $Sim({X}_p, {X}_g)={z}_p ({z}_g)^T$. The illustration on the inference is shown in Fig. \ref{fig:inference}.

The main computational cost in the training comes from the learning parameters of Gaussian posterior, that is, $[\mu_z,\sigma_z]=\phi_r^{Enc} (\phi_r^{x} (x_t), h_{t-1})$ and $[\mu_x,\sigma_x]=\phi_r^{Dec} (\phi_r^{z} (z_t), h_{t-1})$. In our model, we use the pre-trained neural networks \cite{VGG} to extract features from $x_t$ and $z_t$. Thus, $\phi_r^x$ and $\phi_r^z$ are parameterised to be the last fully connected layers of the deep convolutional neural networks \cite{VGG}. Fortunately, the feature extraction $\phi_r^x$ ($\phi_r^z$) can be performed off-line and hence, we consider the complexity regarding $\phi_r^{Enc}$ and $\phi_r^{Dec}$, which are typically a stacked three-layer RNN, and each layer has 1024 LSTM units. As the standard vanilla LSTM is high computational cost because of its complex structure \cite{Bi-lstm}, \eg it has three gates to control the memory cell in addition to the input activation, and thus the computational cost of LSTM is four times larger than a simple RNN with the same dimension of the memory cell. To reduce the computational cost, we compress the output vector of the LSTM by a linear projection referred to as a recurrent projection \cite{lstm-large}. Assume the input vector of $x_t$ is $D$-dimensional, and the dimension of the hidden state $h_t$ is $M$, the sizes of weight matrices of multiplicating the input and the hidden state vectors $W_{\cdot x}$ and $W_{\cdot m}$ are $M\times D$ and $M\times M$, respectively. Therefore, the computational cost of a single LSTM to calculate the $h_t$ from $x_t$ at a time step is approximately $4(D+M)M$. Since the feature dimension from the deep ConvNets \cite{VGG} is 1024, and $M=D=1024$, the number of multiplications becomes $8\times 10^6$. In our model, we compress the output vector $h_t$ into the $R$-dimensional vector via a weight matrix $W_{rh}$ of size $R\times M$. Then the compressed vector is fed as the input to the memory block of the next layer and the hidden state to the gate activations on the same layer at next time step. Thus, the computational cost of this light LSTM becomes $4(D+R)M+RM$. If $M=1024$ and $D=R=256$, the number of multiplications becomes $2\frac{1}{4}\times 10^6$, resulting in 72\% reduction from the vanilla LSTM.

\section{Experiments}\label{sec:exp}

\subsection{Datasets}
\begin{figure*}[t]
\centering
\begin{tabular}{ccc}
\includegraphics[height=3cm,width=5cm]{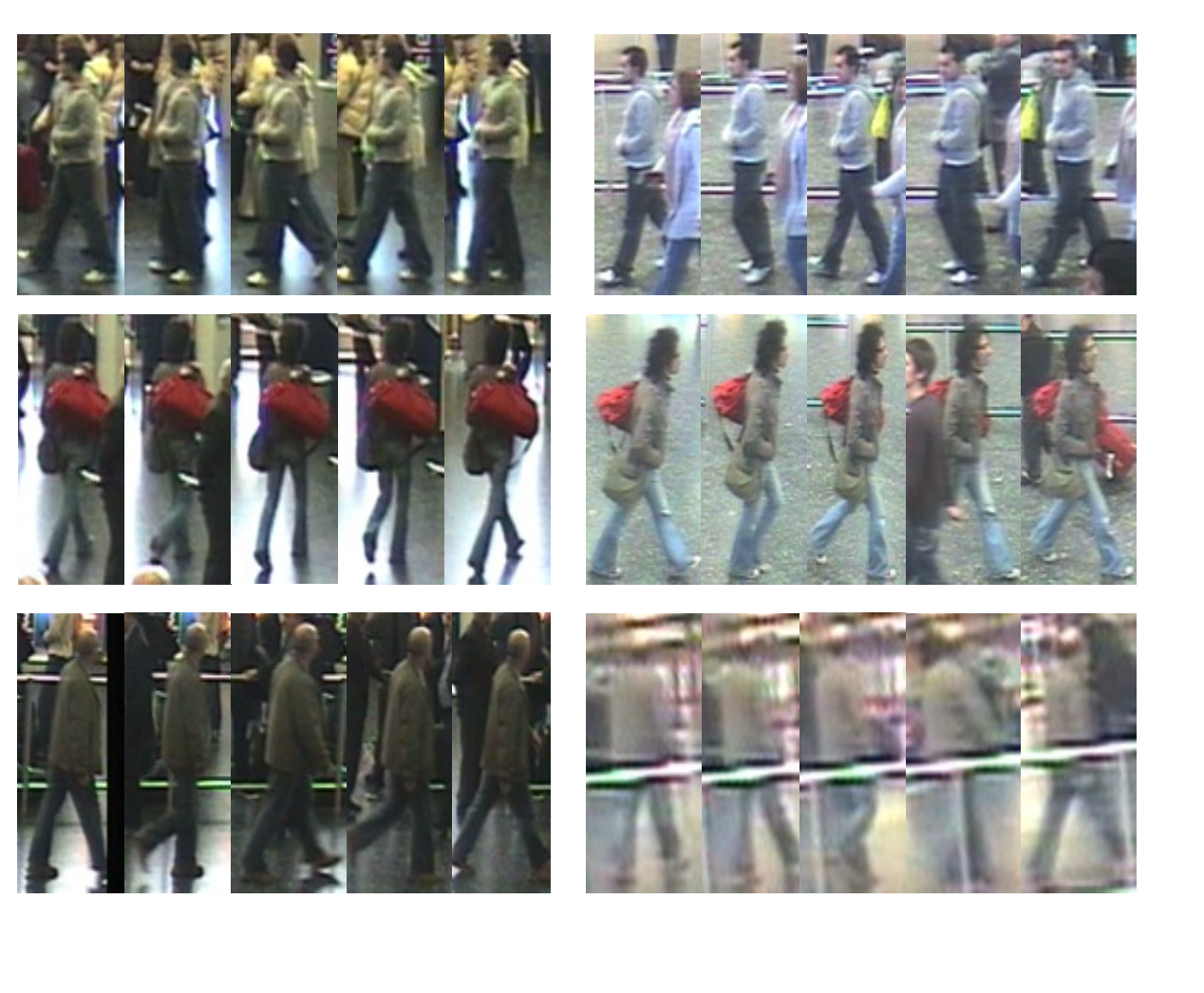}&
\includegraphics[height=3cm,width=5cm]{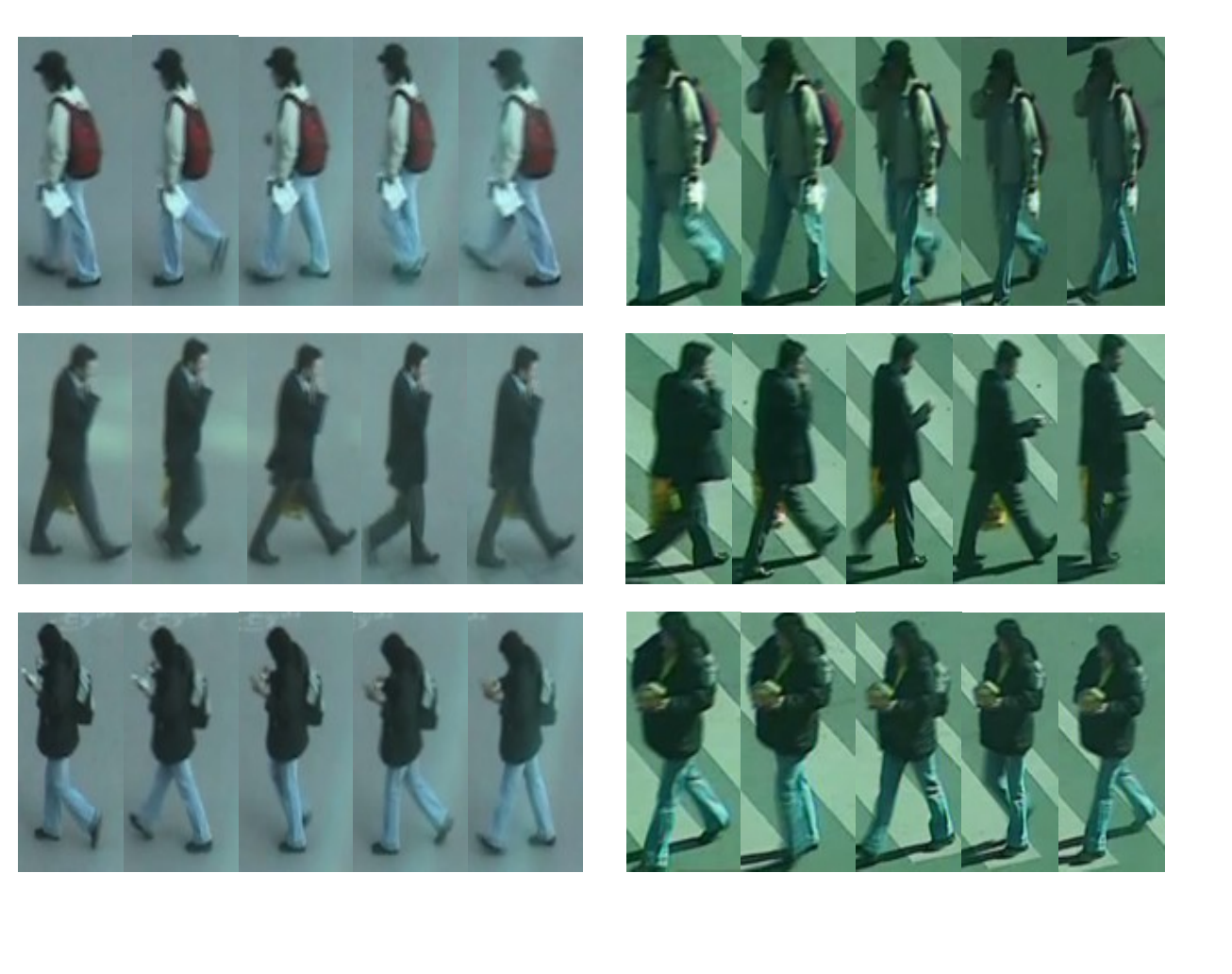}&
\includegraphics[height=3cm,width=5cm]{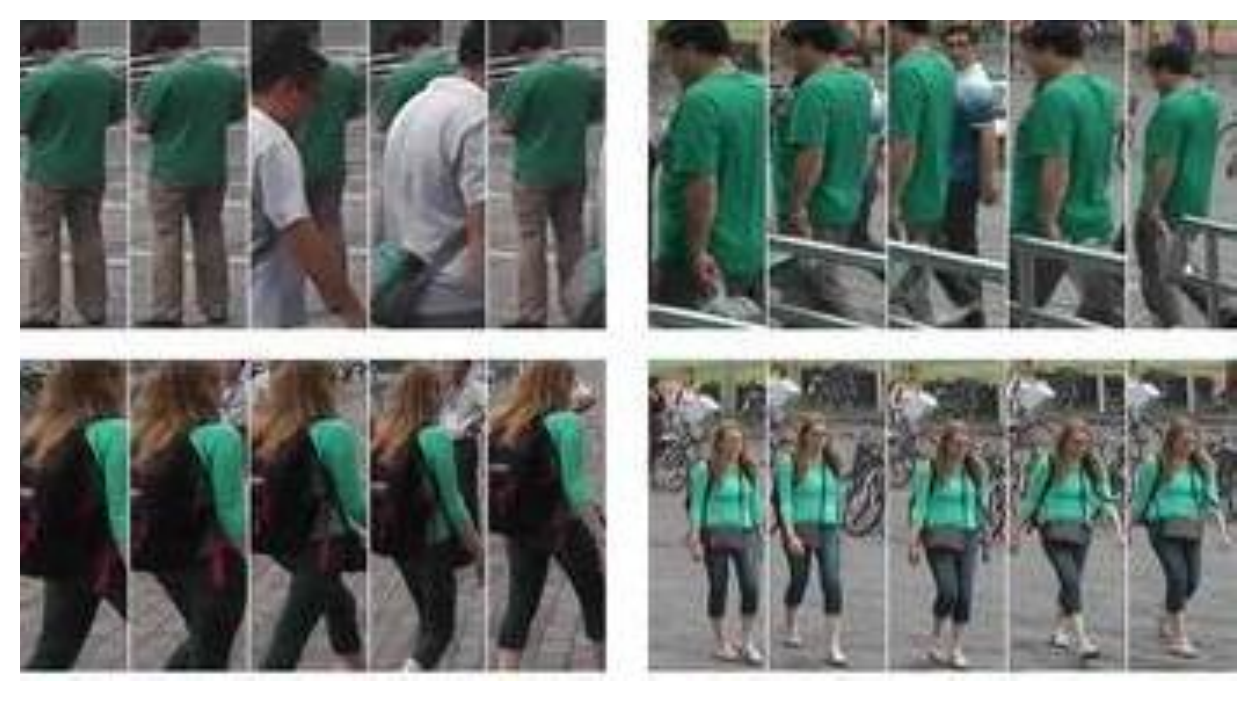}\\
(a) iLIDS-VID & (b) PRID2011 & (c) MARS
\end{tabular}
\caption{Example pairs of image sequence of the same pedestrian in different camera views from three datasets.}
\label{fig:examples}
\end{figure*}

To evaluate the performance of the proposed approach, we conduct extensive experiments on three  image sequence based person re-ID datasets:  iLIDS-VID \cite{VideoRanking}, PRID2011 \cite{PRID2011}, and MARS \cite{MARS}. Example images are shown in Fig. \ref{fig:examples}.

\begin{itemize}
\item The iLIDS-VID dataset consists of 600 image sequences for 300 randomly sampled people, which was created based on two non-overlapping camera views from the i-LIDS multiple camera tracking scenario. The sequences are of varying length, ranging from 23 to 192 images, with an average of 73. This dataset is very challenging due to variations in lighting and viewpoint caused by cross-camera views, similar appearances among people, and cluttered backgrounds.
\item The PRID 2011 dataset includes 400 image sequences for 200 persons from two adjacent camera views. Each sequence is between 5 and 675 frames, with an average of 100. Compared with iLIDS-VID, this dataset was captured in uncrowded outdoor scenes with rare occlusions and clean background. However, the dataset has obvious colour changes and shadows in one of the views.
\item The MARS dataset contains 1,261 pedestrians, and 20,000 video sequences, making it the largest video re-ID dataset. It provides rich motion information by using DPM detector \cite{DetectionPAMI} and GMMCP tracker \cite{GMMCP} for pedestrian detection and tracking. As suggested by \cite{MARS}, the dataset is evenly divided into train and test sets, containing 631 and 630 identities, respectively.
\end{itemize}

\subsection{Experimental Settings}\label{ssec:setting}
We extracted percepts using the convolutional neural net model from \cite{VGG}. To fit the input frame into the model, we resize each frame to be $224\times 224$, and run it through the convnet to produce the RGB percepts. This is implemented by simply re-scaling the largest side of each image to a fixed length, i.e., $224\times 224$ (We use openCV to resize each frame to have 224 width and 224 height). In the case of person re-ID, cropping image has the drawback of potentially excluding critical parts of pedestrians whilst simply resizing an image can still get plausible performance because frames from person re-ID datasets do not have very different aspect ratios. Additionally, we compute flow percepts by extracting flows using the Brox method and train the temporal stream convolutional network as described by \cite{Two-stream}. The optical flow is computed using the off-the-shelf GPU implementation of \cite{Brox-2004} from the OpenCV toolbox. In spite of the fast computation time (0.06s for a pair of frames), it would still impose a bottleneck if done on-the-fly, so we pre-computed the flow before training. To avoid storing the displacement fields as floats, the horizontal and vertical components of the flow were linearly rescaled to a $[0, 255]$ range. We use the 4096-dimensional fc-6 layer as the input representation with RGB and flow percepts.

For the iLIDS-VID and PRID2011 datasets, we randomly split each dataset into two subsets with the same size. One is used for training and the other one for testing. For iLIDS-VID and PRID2011, each of which dataset has only two disjoint cameras and each person has one sequence under each camera. Thus, the sequences from the first camera are used as the probe while the sequences from the second camera are used as the gallery. To constitute the training pairs, each person is regarded as a class, and we only select one probe frame sequence and its gallery correspondence as a training paired input. Please note that for the MARS dataset, we maintain the IDs with the probe and its correspondence and thus we have 625 IDs for training. In the testing on the MARS dataset with more than two cameras, we randomly select one camera as the probe view while a different camera is randomly selected as the gallery. Training was performed by following the Adam optimiser and ran the model for 50 epochs with a learning rate of $10^{-3}$. We set an early stop criteria that the model does not show a decrease in a validation loss for 10 epochs. To set the hyper-parameter such as $\lambda$, we select the parameter by using cross-validation where the labeled paired videos $\{X_p, X_g\}$ in the training are further split into the training set ($S_T$) and the validation set ($S_V$) containing 90\% of the original examples and the rest of 10\%, respectively. We use the labeled set $S_T$ to learn classifiers $G_y(\cdot)$ and $G_d(\cdot)$. The learned classifiers are evaluated on the validation set $S_V$ and parameters are selected corresponding to the classifiers with the lowest validation risk. All our models are implemented on the public Torch \cite{Torch7} platform, and all experiments are conducted on a single NVIDIA GeForce GTX TITAN X GPU with 12 GB memory.

\paragraph{Evaluation Measure} To evaluate the performance of our method and also compare the performance against other methods, we employ the standard Cumulated Matching Characteristics (CMC) curve as our evaluation metric, which indicates the probability of finding the correct match in the top $R$ matches within the ranked gallery. In our experimental results, we report the Rank-$R$ average matching rates, which are obtained by randomly splitting the dataset into training and testing 10 trails and the average result is computed.

\begin{figure}[t]
\centering
\begin{tabular}{c}
\includegraphics[height=5cm,width=7cm]{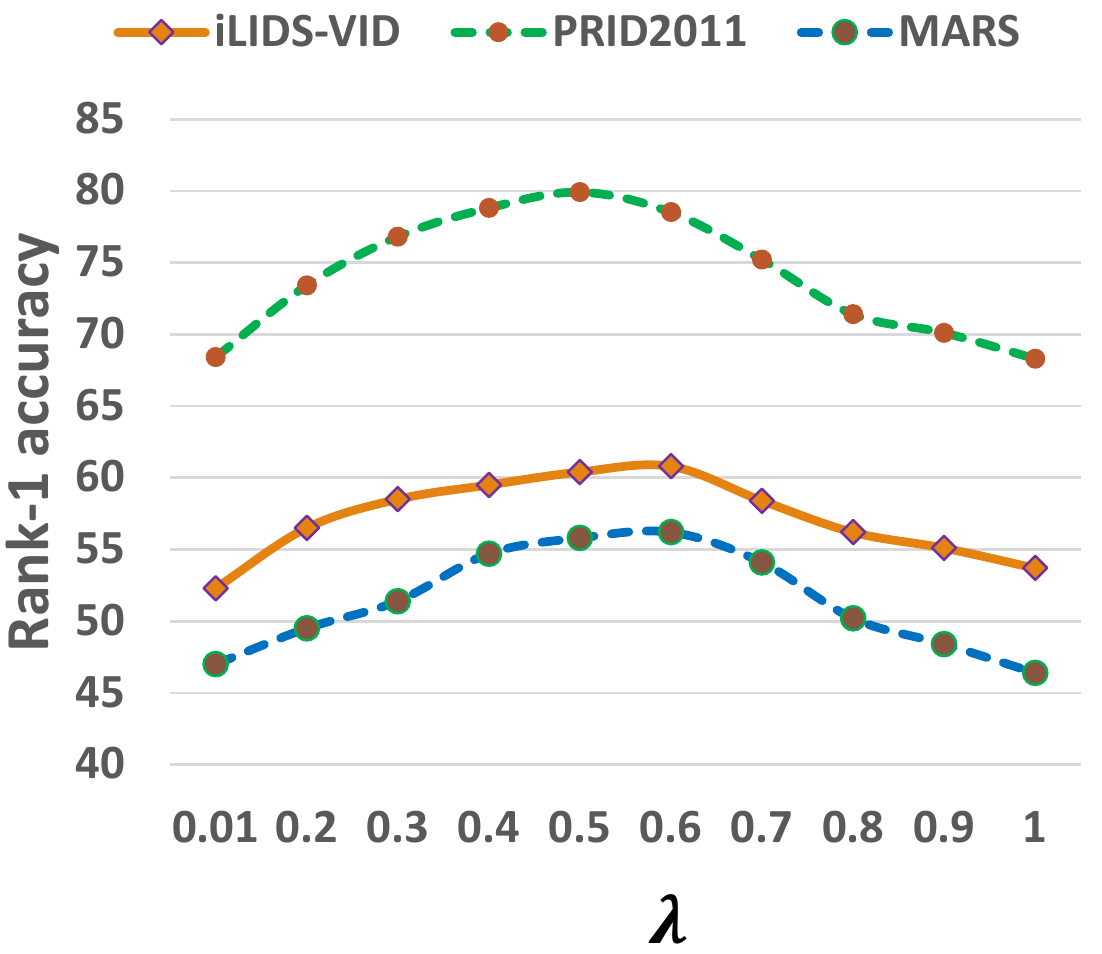}\\
\end{tabular}
\caption{The study on cross-validation risk w.r.t varied values of $\lambda$.}\label{fig:lambda}
\end{figure}

\begin{table}[t]
  \caption{Comparison with different feature learning methods over three datasets. The values of Rank-$R$ indicate the probability of seeking the correct match in the top $R$ matches in the ranked list, and the higher value represents the better performance.} \label{tab:compare-feature}
  {
  \begin{tabular}{|c|c|c|c|c|c|c|}
  \hline
& \multicolumn{2}{c|}{iLIDS-VID} & \multicolumn{2}{c|}{PRID2011}& \multicolumn{2}{c|}{MARS}\\
\cline{2-7}
    Methods  & $R$=1 & $R$=20 & $R$=1 & $R$=20 & $R$=1 & $R$=20\\
  \hline
   V-LSTM \cite{VideoLSTM} & 39.7 & 89.5 & 57.7 & 95.2 & 30.5 & 62.4\\
   VAE \cite{AEV}  & 48.1 & 96.9 & 63.7 & 96.8 & 33.4 & 72.1\\
   R-DANN & 54.0 & 96.9 & 73.8 & 97.7 & 46.4 & 90.2\\
   VRNN   & 51.0 & 95.8 & 69.7 & 97.1 & 41.1 & 88.6\\
   Ours   & $\color{red}\mathbf{60.1}$ & $\color{red}\mathbf{97.9}$ & $\color{red}\mathbf{79.2}$ & $\color{red}\mathbf{98.9}$  & $\color{red}\mathbf{54.2}$  & $\color{red}\mathbf{96.4}$ \\
  \hline
  \end{tabular}
  }
\end{table}

\subsection{Hyper-parameter Selection}

In this section, we set up the hyper-parameter selection on $\lambda$ by using a variant of reverse cross-validation approach \cite{Reverse-valid} to optimise the $\lambda$ in the context of view-invariance adaptation. The evaluation on reverse validation risk associated to varied parameter of $\lambda$ is proceeded as follows. Given the video pairs $[X_p, X_g]$ from the probe ($X_p$) and the gallery view ($X_g$), we split each set into training sets ($X_p^{\ast}$ and $X_g^{\ast}$ respectively, containing 90\% of the original examples) and the validation sets ($X_p^V$ and $X_g^V$ respectively). We use the $X_p^{\ast}$ and $X_g^{\ast}$ to learn a classifier $CLASS(\centerdot)$ to classify each identity across views. Then, we learn a reverse classifier $CLASS(\centerdot)_r$ using the set $\{x, CLASS(x)\}_{x\in X_g^{\ast}}$ and the unclassified part of $X_p^{\ast}$. Finally, the learned reverse classifier $CLASS(\centerdot)_r$ is evaluated on the validation set $X_p^V$, and the classifier $CLASS(\centerdot)$ has a reverse validation risk in the process of addressing cross-view invariance. This process is repeated with multiple values of hyper-parameter $\lambda$ and the selected parameter is the one corresponding to the classifier with the lowest reverse validation risk. As shown in Fig. \ref{fig:lambda}, the adaptation parameter $\lambda$ is chosen among 9 values between $10^{-2}$ and 1 on a logarithmic scale, and the optimal value of $\lambda$ is chosen to be 0.6 when the method has the lowest validation risk.

\subsection{Ablation Studies}

\begin{figure}[t]
\centering
\begin{tabular}{c}
\includegraphics[height=4cm,width=9cm]{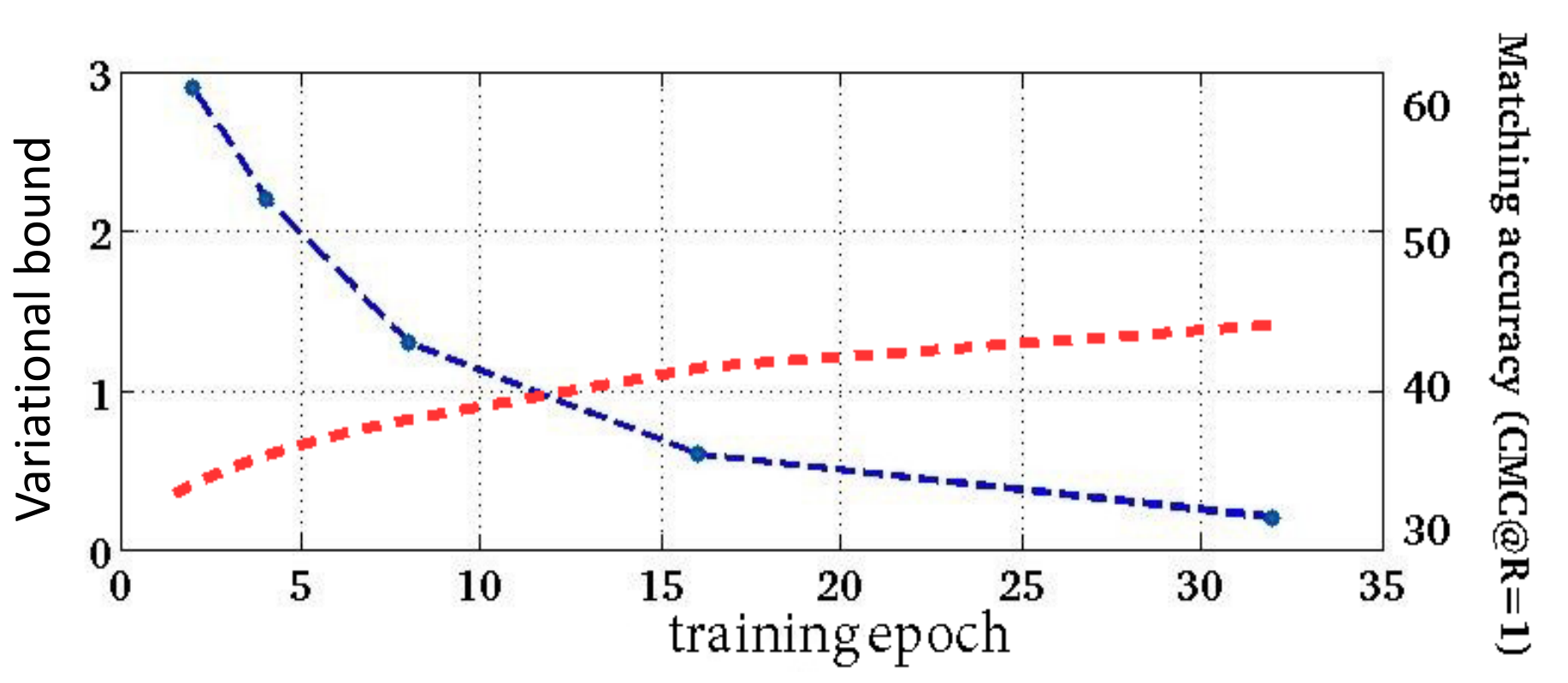}\\
\end{tabular}
\caption{The effect of cross-camera feature distribution in matching accuracy of video-based person re-ID. The blue line represents the variational bound w.r.t training epochs while the red line denotes the matching accuracy w.r.t the same training epochs.}\label{fig:distribution_match}
\end{figure}

In this section, we conduct extensive experimental analysis to answer the following questions: (a) How does our method perform compared to other feature learning models for videos? (b) How do we show that the learned temporal latent features are effective in matching persons against view changes? (c) How is the performance of our method affected with respect to varied input lengths of videos? (d) How large the computational complexity is as opposed to existing deep recurrent models?

\paragraph{Comparison with Other Feature Learning Models for Videos} To evaluate the effectiveness of latent video representations learned by our model, we compare with state-of-the-art deep generative models:
\begin{itemize}
  \item R-DANN: The domain adversarial neural networks \cite{DANN} is a deep domain adaptation model which uses two components to create domain-invariant representations: a feature extractor that produces the data's latent representations, and an adversarial domain labeller that helps the feature extractor produce features domain-invariant. In our experiment, the feature exactor is a RNN (LSTM) with fc6 as the input representation of data where each LSTM has 2048 units.
  \item V-LSTM \cite{VideoLSTM}: This method uses LSTM networks to learn representations of video sequences. The state of the encoder of LSTM after the last input has been read is the representation of the input video, and similar to R-DANN, each LSTM has 2048 units.
  \item VAE \cite{AEV}: It offers high-level latent random variables to model the variability observed in the data with a combination of highly flexible non-linear mapping between the latent random state and the observed output and effective approximate inference.
  \item VRNN \cite{VRNNs}: It extends the VAE into a recurrent framework for modelling the high-dimensional sequences.
\end{itemize}

The comparison results are given in Table \ref{tab:compare-feature}. The proposed method is superior in all datasets with large improvements over V-LSTM \cite{VideoLSTM}. The lower recognition accuracy obtained by LSTM model is possibly due to the difficulties of training with large visual variations and cross-view divergence. VAE \cite{AEV} performs better than V-LSTM \cite{VideoLSTM} by introducing latent random variables which are effective in learning the mapping between videos and latent representations. R-DANN \cite{DANN} outperforms VRNN \cite{VRNNs} by performing domain-invariance training, however, R-DANN is not focusing on the underlying temporal dependencies. In contrast, our approach explicitly addresses the cross-view feature distribution changes by adversarial training on the latent variables produced by VRNNs.

\paragraph{View-invariance Study} The second experiment is to demonstrate the property of our features in view-invariance. To this end, we evaluate the matching performance of our method against the difference between feature distributions. Fig.\ref{fig:distribution_match} with dual Y-axis shows two evaluations: The red line indicates the performance of our method for video matching by reducing the parameter distributions across camera views as the training epochs proceeds. The blue line represents the difference between feature distributions under two camera views measured by the KL-divergence. In this experiment, we use the testing data from the iLIDS-VID dataset, and keep the representations by increasing the number of training epochs. We can see an improvement in the matching accuracy as more training epochs are proceeded, which is aligned with the decrease in feature distribution divergence as measured by the variational bound of KL-divergence.

\begin{figure*}[t]
\begin{tabular}{ccc}
\includegraphics[height=4cm]{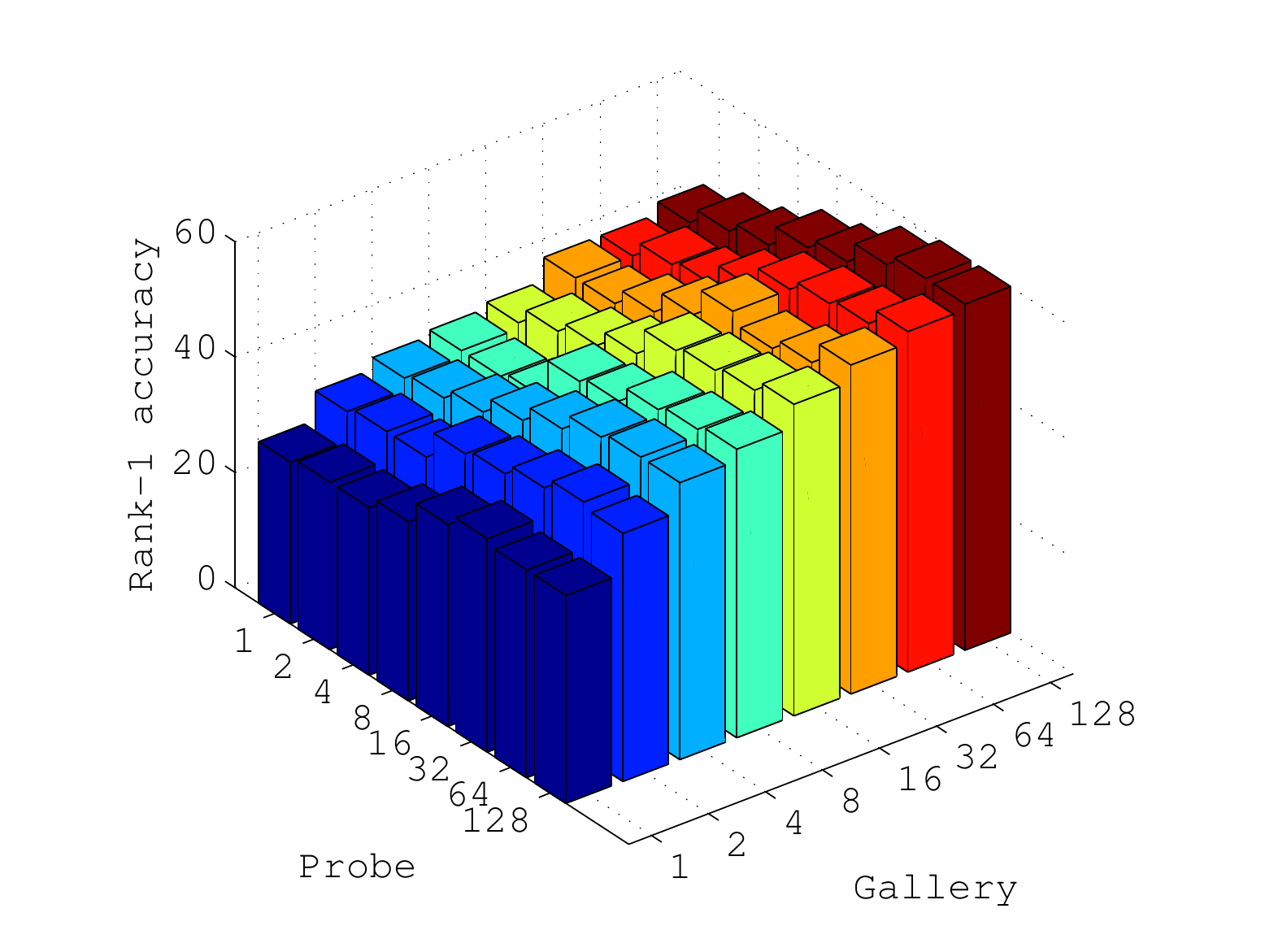} &
\includegraphics[height=4cm]{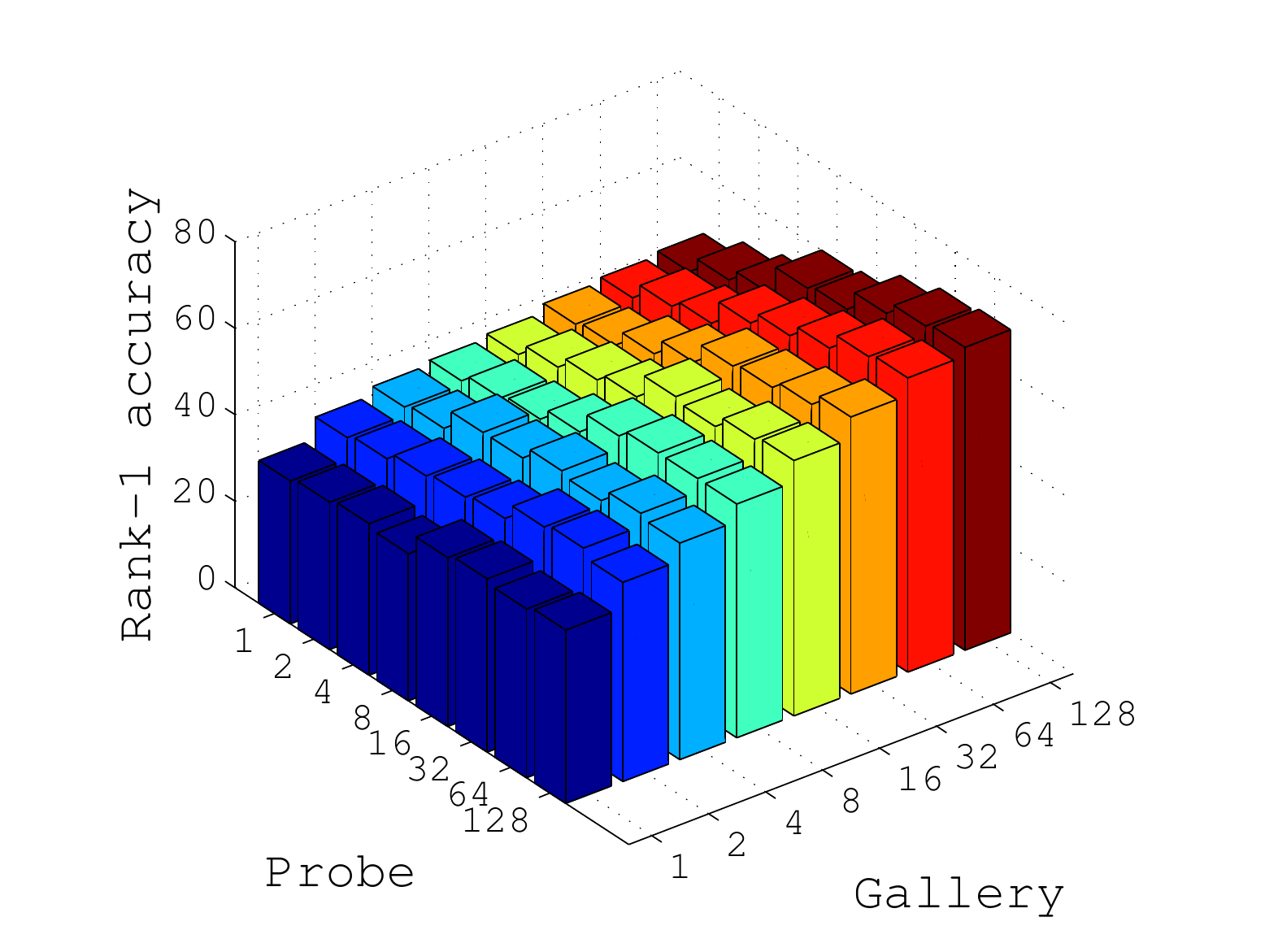} &
\includegraphics[height=4cm]{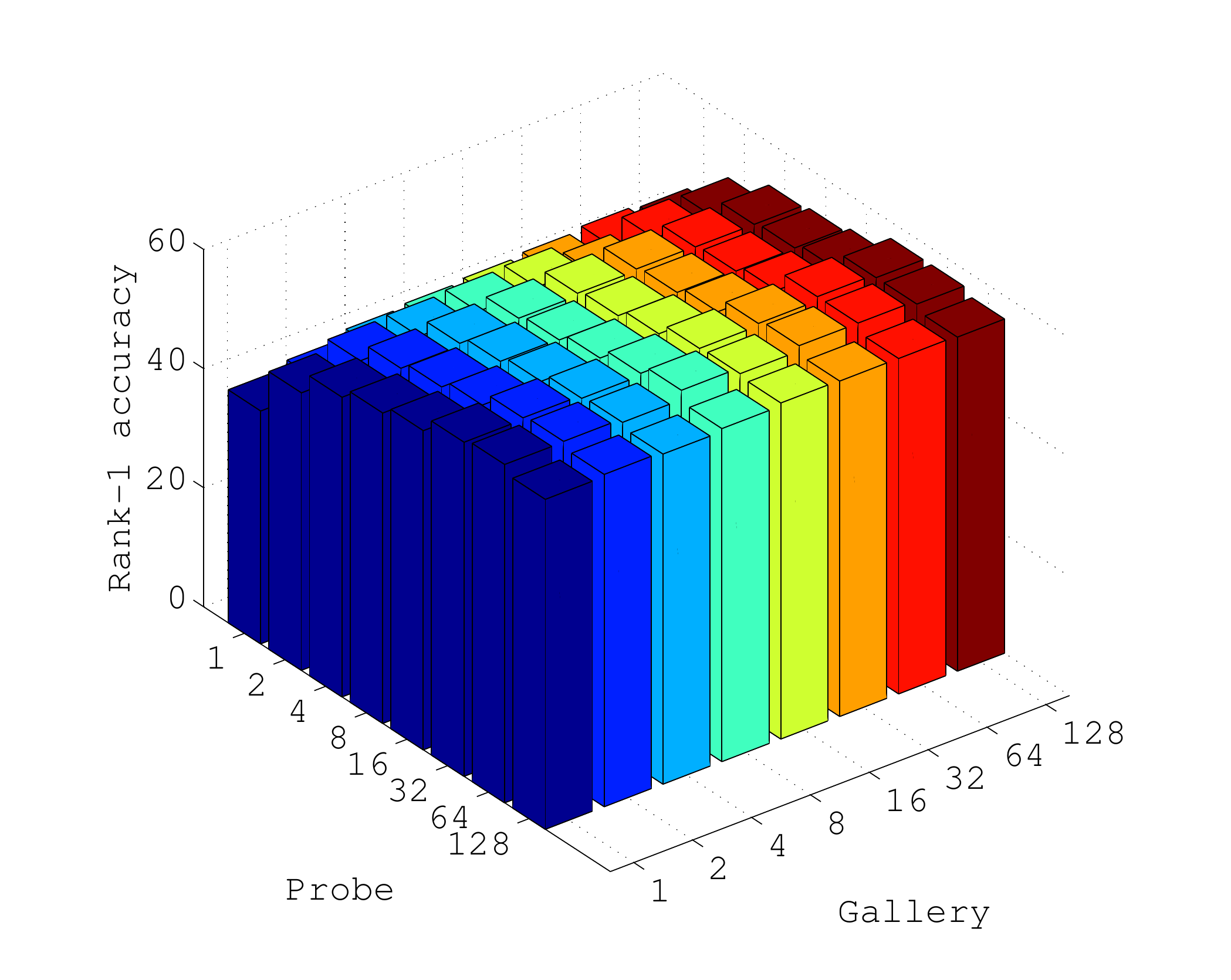} \\
(a) iLIDS-VID & (b) PRID2011 & (c) MARS
\end{tabular}
\caption{Rank-1 matching rate as varied length of the probe and gallery sequences.}
\label{fig:varied_length}
\end{figure*}

\paragraph{Variable Length} The third study is to investigate how the performance varies against the length of the probe and gallery sequences in testing. Evaluations are conducted on three datasets, and the lengths of the probe and gallery sequences are varied between 1 and 128 frames in step of power of two. Results are shown in Fig.\ref{fig:varied_length} where a bar matrix shows the rank-1 re-identification matching rate as a function of the probe and gallery sequence lengths. We can see that increasing the length in either probe or gallery can increase the matching accuracy. Also longer gallery sequences can bring about more performance gain than longer probe sequences.

\paragraph{Computational Cost Analysis} The final study is to analyse the computational cost of our model as opposed to existing deep recurrent models. We consider the vanilla LSTM \cite{Bi-lstm} with three layers as comparison. The vanilla LSTM and the light LSTM in our model are trained with 1024-dimensional memory cells and the light LSTM has an additional 256-dimensional recurrent projection layer. The comparison results are shown in Table \ref{tab:computational-cost}. It can be seen that the LSTM with projection outperforms the vanilla LSTM with the number of weight parameters one third of the vanilla LSTM.

\begin{table}[t]
  \caption{Comparison of LSTMs complexity and performance of Rank-1 over two datasets.} \label{tab:computational-cost}
  {
  \begin{tabular}{|c|c|c|c|}
  \hline
&  & \multicolumn{2}{c|}{Rank-1} \\
\cline{3-4}
    Models  & Number of Weights &  iLIDS-VD & PRID2011\\
  \hline
   vanilla LSTM \cite{Bi-lstm} & 22,595,584 & 59.6 & 79.0 \\
   Ours   & 7,548,160 &  60.1 &  79.3\\
  \hline
  \end{tabular}
  }
\end{table}

\subsection{Impact of Adversarial Learning}
\begin{figure}[t]
\centering
\begin{tabular}{c}
\includegraphics[height=5cm,width=6cm]{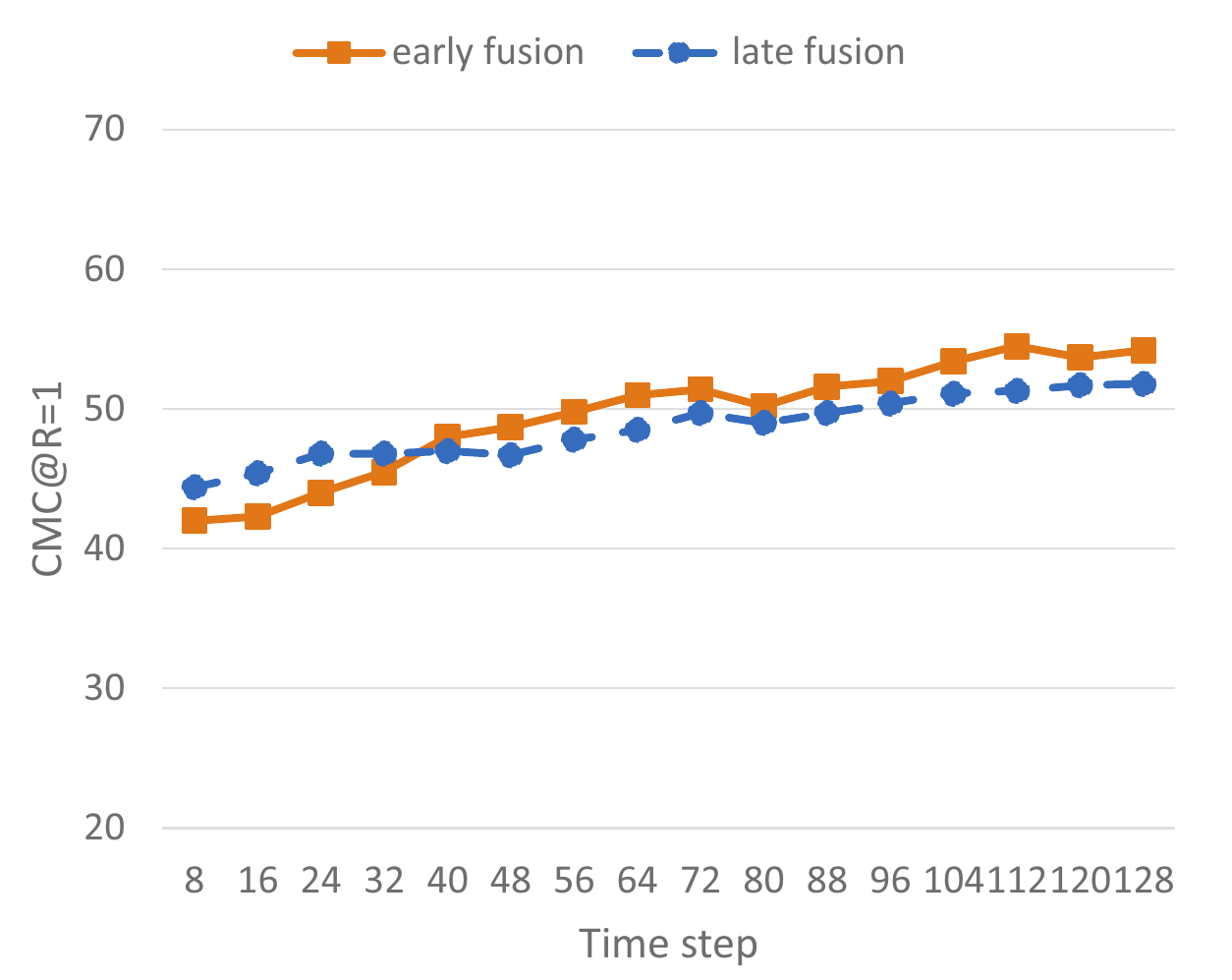}\\
\end{tabular}
\caption{The impact of adversarial learning on MARS dataset.}\label{fig:impact-adv}
\end{figure}

It is an interesting question to know what is the right time to perform the adversarial operation? It should be at every time step or the last time step of a latent representation learning. If the adversarial training is performed at every time step, namely early fusion, the network learns to produce the view-invariant representations conditioned on the subsets of the input sequence $x_{\leq T}$. On the other hand, a late adversarial learning at the last time-step (i.e., late fusion) might be incapable of eliminating feature variations during the representation learning. To study the impact of adversarial learning on the optimal view-invariant representations, we empirically test the two different strategies by assessing the CMC values at $R=1$ computed at each time-step. The comparison results on MARS dataset is shown in Fig.\ref{fig:impact-adv}. The results indicate that progressively performing adversarial learning at each time-step on the learned features lead to more optimal representations.

\subsection{Comparison with Sate-of-the-art Approaches}

In this section, we compare our method with state-of-the-art video-based person re-ID approaches. We consider the following competitors:
\begin{itemize}
\item XQDA \cite{LOMOMetric}: Cross-View Quadratic Discriminant Analysis is a static appearance feature based supervised approach that learns simultaneously a discriminant low dimensional subspace and a QDA metric.
\item eSDC \cite{eSDC}: An effective unsupervised spatial appearance based method, which is able to learn localised appearance saliency statistics for measuring local patch importance.
\item SDALF \cite{Farenzena2010Person}: A classic hand-crafted visual appearance feature for re-ID purpose.
\item ISR \cite{Giuseppe2015PAMI}: A weighted dictionary learning based algorithm that iteratively extends sparse discriminative classifiers.
\item HOG3D+DVR \cite{VideoRanking}: It uses a motion feature HOG3D \cite{HOG3D} for representing video slices, and performs discriminative video ranking.
\item STFV3D \cite{VideoPerson}: A low-level feature-based Fisher vector learning and extraction method which is applied to spatially and temporally aligned video fragments.
\item RCN \cite{RCNRe-id}: A contemporary deep learning approach that incorporates CNN underlying LSTMs to learn video-level features for person re-ID.
\item TS-DTW \cite{Video-person-matching}: A unsupervised method to extract space-time person representation by encoding multiple granularities of spatiotemporal dynamics in form of time series.
\item TDL \cite{Top-push}: A top-push distance learning (TDL) model incorporating a top-push constraint to quantify ambiguous video representation for video-based person re-ID.
\item RFA-net \cite{RFA-net}: It uses the Long-Short Term Memory (LSTM) network to aggregate frame-wise person features in a recurrent manner.
\item $SI^2DL$ \cite{Video-person-ijcai16}: A set-based supervised distance learning model which aims to learn a pair of intra-video and inter-video distance metrics.
\item What-and-where \cite{Wu-TMM}: An end-to-end Siamese-like network to match videos of person re-ID by attending to distinct local regions.
\item EUG \cite{EUG}: It exploits the unlabelled tracklets to update the CNNs by stepwise learning.
\end{itemize}

\begin{figure}[t]
\centering
\begin{tabular}{c}
\includegraphics[height=4.5cm]{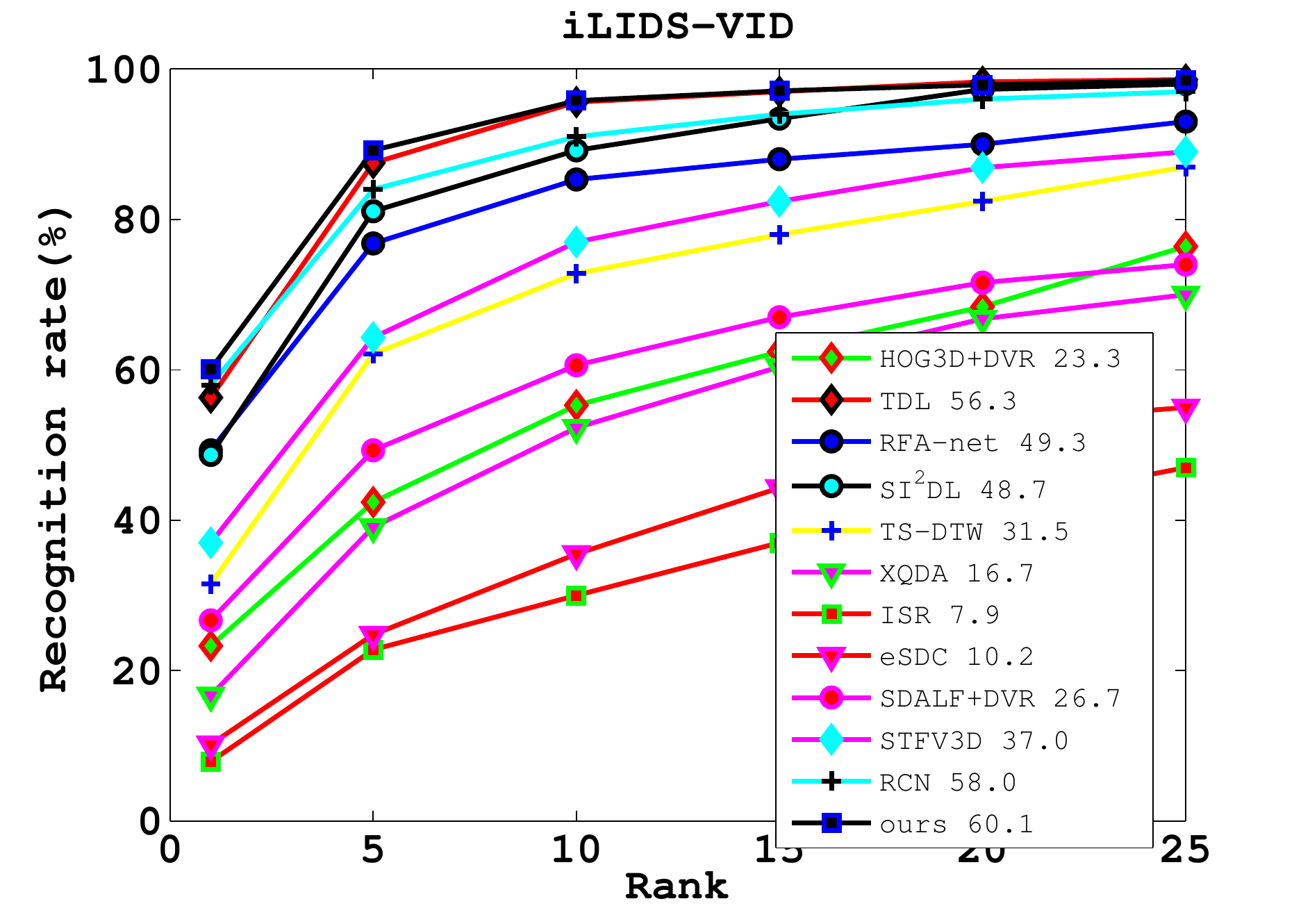} \\
 (a) iLIDS-VID \\
\includegraphics[height=4.5cm]{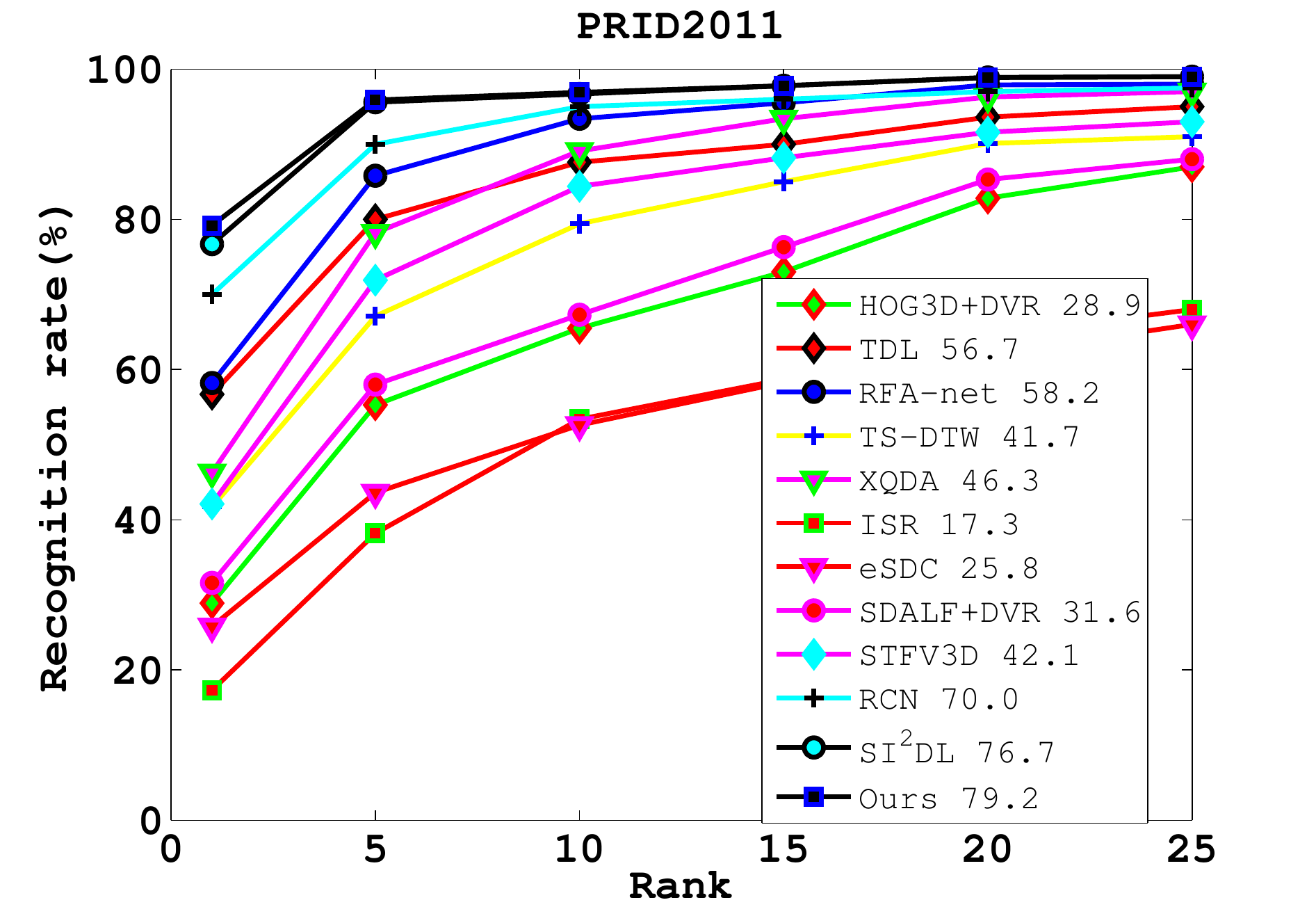}\\
 (b) PRID2011
\end{tabular}
\caption{CMC curve on the iLIDS-VID and PRID2011 datasets. Rank-1 matching rate is marked after the name of each approach.}\label{fig:match_rate}
\end{figure}

The comparison results on three datasets are given in Table \ref{tab:cmc_value} and Fig.\ref{fig:match_rate}. It can be seen that our method has large improvement gains compared against multi-shot methods including XQDA \cite{LOMOMetric}, ISR \cite{Giuseppe2015PAMI}, and eSDC \cite{eSDC}, in which temporal dependency across frames are not considered. Compared with 3D feature based methods such as HOG3D \cite{HOG3D}+DVR \cite{VideoRanking}, SDALF\cite{Farenzena2010Person}+DVR \cite{VideoRanking}, and STFV3D \cite{VideoPerson}, our method has a high matching rate by learning hidden representations which are able to faithfully describe person videos with variations. Competing methods of RCN \cite{RCNRe-id} and RFA-net \cite{RFA-net} are based on RNN structure and use deterministic functions to compute transitions from inputs into hidden states, which cannot capture data variability to be addressed in cross-view video matching, and thus leads to inferior performance to our approach. In contrast, our approach explicitly address cross-view difference by adversarial learning, and the learned video features are more effective in person re-ID. For example, on iLIDS-VID dataset, RCN \cite{RCNRe-id} attains 58.0\% at rank-1 matching rate while the proposed method achieves 60.1\% at rank-1 value. On PRID-2011 dataset, our method outperforms $SI^2DL$ \cite{Video-person-ijcai16} by 2.5\% at rank-1 while using few labeled training pairs. One thing to be noticed is that our method performs secondary to the method What-and-where \cite{Wu-TMM} in rank-1 and rank-5 on the MARS dataset. The possible reason is What-and-where \cite{Wu-TMM} is based on supervised learning to match persons with annotations on body parts. In contrast, our method does not require any region annotations. In the combination with KissME \cite{Kostinger2012Large}, Ours+KissME \cite{Kostinger2012Large} has further improvement on recognition values on all datasets. For instance, Ours+KissME \cite{Kostinger2012Large} achieves the best results on rank-1 (64.6\% and 84.2\%) on the iLIDS-VIDS and PRID011 datasets, and the best mAP (52.1\%) on the MARS dataset. This allows us to potentially improve the performance in combining with metric learning algorithms.

\begin{table*}[t]
\centering
  \caption{Rank-1, -5, -10, -20 recognition rate of different methods on the iLIDS-VID, PRID2011 and MARS datasets. }  \label{tab:cmc_value}
 {
  \begin{tabular}{|c|c|c|c|c|c|c|c|c|c|c|c|c|c|c|}
  \hline
& \multicolumn{4}{c|}{iLIDS-VID} & \multicolumn{4}{c|}{PRID2011}& \multicolumn{5}{c|}{MARS}\\
\cline{2-14}
    Methods  &  $R=1$  & R=5 & R=10 & R=20 & R=1 & R=5 & R=10  & R=20 &  R=1  & R=5 & R=10 & R=20 & mAP\\
  \hline
   XQDA \cite{LOMOMetric} & 16.7 & 39.1& 52.3 & 66.8  & 46.3 & 78.2& 89.1& 96.3 & 30.7 & 46.6 & 53.5 & 60.9 & 16.4\\
   ISR \cite{Giuseppe2015PAMI}  & 7.9 & 22.8  & 30.0 & 41.8  & 17.3 &38.2 & 53.4 & 64.5 & - & -&-  &- &-\\
   eSDC \cite{eSDC}  & 10.2 &24.8 & 35.5 & 52.9 & 25.8 & 43.6 & 52.6 & 62.0 &-  &- &- &- &-\\
  HOG3D \cite{HOG3D}+DVR  \cite{VideoRanking} & 23.3 & 42.4 & 55.3 & 68.4 & 28.9 & 55.3 & 65.5 & 82.8 & 12.4 &33.2 & 54.7 &71.8 &-\\
  SDALF \cite{Farenzena2010Person} +DVR \cite{VideoRanking} & 26.7  & 49.3  & 60.6 & 71.6 & 31.6 & 58.0 & 67.3 & 85.3& 4.1& 12.3& 20.2 & 25.1 &1.8\\
  STFV3D  \cite{VideoPerson} & 37.0  & 64.3 & 77.0 & 86.9 & 42.1 & 71.9& 84.4 & 91.6 & - & -&- &- &-\\
  RCN  \cite{RCNRe-id} & 58.0  & 84.0 & 91.0& 96.0& 70.0 &90.0 & 95.0 & 97.0 & - & -&- &- &-\\
  TS-DTW \cite{Video-person-matching}  & 31.5 & 62.1 & 72.8 & 82.4 & 41.7 & 67.1& 79.4 & 90.1 & - & -&- &- &-\\
  TDL\cite{Top-push}  & 56.3 & 87.6 & 95.6 & $\mathbf{98.3}$ & 56.7 & 80.0 & 87.6 & 93.6 & - & -&- &- &-\\
  RFA-net \cite{RFA-net}  & 49.3 & 76.8 & 85.3 & 90.0 & 58.2 & 85.8 & 93.4 & 97.9 & - & -&- &- &-\\
  $SI^2DL$ \cite{Video-person-ijcai16} & 48.7 & 81.1 & 89.2 & 97.3 & 76.7 & 95.6& 96.7 &98.9  & - & -&- &-&-\\
  CNN+KissME+MQ \cite{MARS} & 48.8 & 75.6 & 84.5 & 92.6 & 69.9  & 90.6 & 96.5 & 98.2 & 68.3 & 82.6 & 86.0 & 89.4 &49.3\\
  EUG \cite{EUG} & - & -& - & -& - &- &-  &-  & 62.6 &74.9 & 79.7& 82.6 & 42.4\\
  What-and-where \cite{Wu-TMM} & 61.2 & 80.7 & 90.3 & 97.3 & 74.8 & 92.6 & $\mathbf{97.7}$ & 98.6 & $\mathbf{69.7}$ & $\mathbf{83.4}$ & 88.3 & $\mathbf{96.6}$ & 50.5\\
\hline
   Ours & 60.1 & $\mathbf{89.2}$ & $\mathbf{95.8}$ &  97.9 & $\mathbf{79.2}$ &  $\mathbf{95.9}$ & 97.9 & $\mathbf{98.9}$ & 54.6 & 76.5 & $\mathbf{89.7}$ & 96.4 & $\mathbf{50.7}$\\
  Ours+KissME \cite{Kostinger2012Large} & $\mathbf{64.6}$  & $\mathbf{90.2}$ & $\mathbf{95.9}$ &  97.9 & $\mathbf{84.2}$ &  $\mathbf{96.9}$ & $\mathbf{97.7}$ & $\mathbf{98.9}$ & 61.2 & 79.5 & $\mathbf{90.7}$ & $\mathbf{96.9}$ & $\mathbf{52.1}$\\
  \hline
  \end{tabular}
  }
\end{table*}

\subsection{Comparison with Other Few-Shot Learning Methods}

\begin{figure}[t]
\centering
\begin{tabular}{c}
\includegraphics[height=5cm,width=8cm]{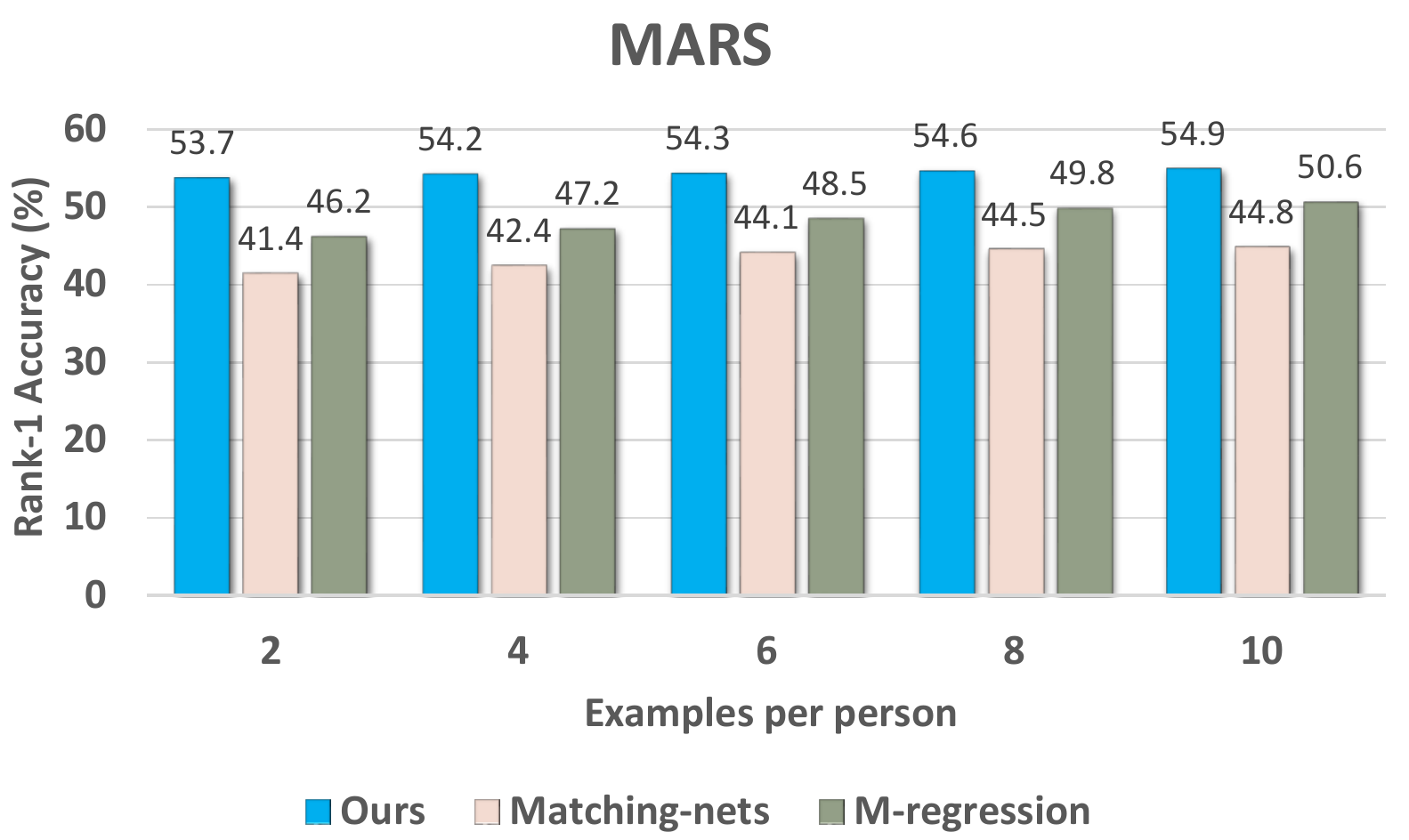}\\
\end{tabular}
\caption{The comparison with two recent few-shot learning methods.}\label{fig:compare-few-shot}
\end{figure}

We also compared to two recently proposed few-shot learning methods: matching networks \cite{Match-nets} and model regression \cite{Model-regress}. The matching networks propose a nearest neighbor approach that trains an embedding end-to-end for the task of few-shot learning. Model regression trains a small MLP to regress from the classifier trained on a small dataset to the classifier trained on the full dataset. Both of the two techniques are high-capacity in learning from few examples and facilitates the recognition in the small sample size regime on a broad range of tasks, including domain adaptation and fine-grained recognition. Comparison results are shown in Fig. \ref{fig:compare-few-shot}. In terms of the overall performance, our method outperforms the two competitors constantly over the MARS dataset. Matching networks exhibit similar performance to our method, however, matching networks are based on nearest neighbors and use the entire training set in memory, and thus they are more expensive in testing time compared with our method and model regressors. Please note that we do not perform the comparison on iLIDS-VID and PRID2011 datasets because these two datasets have only two cross-view sequences regarding  each person.

\section{Conclusion and Future Work}
We propose a novel few-shot deep adversarial model with a latent variational structure to learn deep temporal representations for video-based person re-identification with few labeled training examples. The proposed method is based on variational recurrent neural network which provides a family of latent variables to describe input videos with temporal relationships. To eliminate feature distribution divergence caused by view changes, the learned latent features are adversarially trained through a cross-view verification loss to make them view-invariant. The proposed method requires few labeled training examples while generic enough and scalable to larger networked cameras. In future work, we would explore one-shot or zero shot learning with memories to augment the generalisation of the model.

\section*{Acknowledgement}

Yang Wang was supported by National Natural Science Foundation of China, under Grant No 61806035. Meng Wang was supported by National Natural Science Foundation of China, under Grant No 61432019, 61732008, and 61725203.

\ifCLASSOPTIONcaptionsoff
  \newpage
\fi



\bibliographystyle{IEEEtran}\small
\bibliography{allbib}

\end{document}